\documentclass{esannV2}
\usepackage[dvips]{graphicx}
\usepackage[latin1]{inputenc}
\usepackage{amssymb,amsmath,array,booktabs}
\usepackage{xcolor}
\usepackage{url}

\begin{document}
\title{The Structure-Content Trade-off in Knowledge Graph Retrieval}

\author{Valentin Six and Ga\"{e}l de Chalendar and Evan Dufraisse
\vspace{.3cm}\\
Universit\'{e} Paris-Saclay - CEA, Palaiseau - France}

\maketitle

\begin{abstract}
Large Language Models (LLMs) increasingly rely on knowledge graphs for factual reasoning, yet how retrieval design shapes their performance remains unclear. We examine how question decomposition changes the retrieved subgraph's content and structure. Using a hybrid retrieval function that controls the importance of initial question and subquestions, we show that subquestion-based retrieval improves content precision, but yields disjoint subgraphs, while question-based retrieval maintains structure at the cost of relevance. Optimal performance arises between these extremes, revealing that balancing retrieval content and structure is key to effective LLM reasoning over structured knowledge.
\end{abstract}

\section{Introduction}

Large Language Models (LLMs) have achieved remarkable success across many natural language tasks but often struggle with factual consistency and multi-hop reasoning \cite{huang2025survey}. To mitigate these issues, Knowledge Graph Question Answering (KGQA) systems ground LLMs in structured, verifiable knowledge \cite{yasunaga2021qa}, providing explicit reasoning paths over factual relations.

A key challenge in KGQA lies in how relevant regions of the graph are retrieved for a given question. An early idea was to decompose complex questions into simpler subquestions \cite{fu-etal-2021-decomposing-complex, perez2020unsupervised}, improving interpretability and modular reasoning. However, how decomposition reshapes the retrieved subgraph, both in terms of its content and structure, remains largely unexplored. This raises important questions such as: how does question decomposition affect what we retrieve from the graph, and how does decomposition affect downstream LLM reasoning?

We introduce a parametric hybrid retrieval framework that interpolates between subquestion and question-focused retrieval. By adjusting a single parameter, we allow systematic exploration of how retrieval focus shapes both the content and structure of retrieved subgraphs, and how these factors interact to influence LLM-based reasoning. We evaluate retrieval quality and downstream performance on two multi-hop QA datasets, CWQ \cite{talmor2018web} and WebQSP \cite{yih2016value}.

Our contributions are threefold: \textbf{(1)} We introduce a parametric framework that enables systematic study of query decomposition effects, \textbf{(2)} we highlight the existence of a structure-content trade-off induced by query decomposition, and \textbf{(3)} we connect this intrinsic trade-off to QA accuracy, demonstrating how retrieval structure governs downstream reasoning accuracy. Together, these results highlight that effective LLM reasoning over structured knowledge requires balancing what is retrieved (content) with how it is connected (structure). \footnote{This paper has been or will be submitted to ESANN 2026.}

\section{Related Work}

Recent advances in KGQA have explored how retrieval structure shapes LLM reasoning. In G-Retriever \cite{he2024g}, they formulate the retrieval process as an optimization problem, forcing the retrieved subgraph to be a single connected-component that maximizes relevance of nodes and edges while minimizing the size of the graph. Previous work also highlights a trade-off between content relevance and graph size \cite{dai2025large}: increasing node coverage often introduces noise, whereas smaller graphs can leave out highly relevant nodes. These observations highlight the importance of subgraph structure when performing retrieval. Moreover, recent state-of-the-art RAG approaches have been using query decomposition \cite{chan2024rq, li2024framework} to improve reasoning by splitting complex queries into simpler subquestions. However, these approaches focus on raw task performance without understanding how query decomposition truly affects the retrieved subgraph. Our work unifies these perspectives by studying how query decomposition shapes the subgraph content and structure, revealing how this balance governs LLM reasoning over knowledge graphs.

\section{Hybrid Retrieval for Graph Construction}

Given a complex question $Q$ and a knowledge graph $G = (V, E)$, the goal is to retrieve a subgraph $G^* \subset G$ that supports answering $Q$. The question is decomposed into atomic subquestions $\{q_1,\dots, q_n\}$, each isolating a semantic aspect of the reasoning chain. For example, the question \textit{'Who directed the film starring Tom Hanks in 1994?'} decomposes into subquestions about Tom Hanks's filmography and film directors. To study the role of both the initial question $Q$ and the corresponding subquestions $\{q_1,\dots, q_n\}$ , we introduce a new similarity function that allows us to control the influence of both sides:
\begin{equation}
\text{sim}_\alpha(z, z_q, z_Q) = (1-\alpha) \cdot \cos(z, z_Q) + \alpha\cdot\cos(z, z_q)
\end{equation}
We encode each subquestion $q$ as $z_q$, and the initial question $Q$ as $z_Q$. Each node and edge of the graph is characterized by their textual entity label, which is encoded as $z$. The coefficient $\alpha \in [0,1]$ controls retrieval focus: $\alpha=0$ emphasizes the initial question $Q$, and $\alpha=1$ the subquestions $\{q_1,\dots, q_n\}$. For each $q$, we retrieve $k$ nodes and $k$ edges with the highest similarity score to form the subgraph $G_q$:
\begin{equation}
G_q = \text{top-}k\,[\,\text{sim}_\alpha(z, z_q, z_Q)\,]
\end{equation}
We retrieve nodes and edges independently to ensure relation relevance is not limited by entity selection. Following \cite{he2024g}, we extract a single connected subgraph per sub using a Prize-Collecting Steiner Tree, allowing minimal, high-value graphs with selective node/edge pruning. We finally merge  the resulting subgraphs into $G^*$, using structural union without duplication:
\begin{equation}
    G^* = \bigcup_{q \in \{q_1, \dots, q_n\}} G_q
\end{equation}
Note that although each intermediate $G_q$ is connected, the union $G^*$ is not necessarily connected. 

This formulation enables us to explicitly vary retrieval focus and observe how it affects both the structure of the resulting subgraph, and the relevance of retrieved entities. In the next section, we evaluate how $\alpha$ shapes this structure-content balance (\textbf{RQ1}) and how this balance influences reasoning performance (\textbf{RQ2}).

\section{Experimental Evaluation}

To measure the influence of question and subquestions, we vary $\alpha$ from 0 to 1 with 0.1 increments. We set $k=5$ for the retrieval stage. We encode questions and subquestions using a sentence encoder \cite{sentence-transformers_all-roberta-large-v1_2025}. Given the format of entity labels, we choose to encode node and edges textual attributes using the same model. For question decomposition\footnote{See \url{https://github.com/cea-list-lasti/decomp-reasoning} for corresponding code and prompt templates.}, we use DeepSeek-R1-Distill-Qwen. We manually evaluate the quality of generated subquestions with different LLM sizes in Table~\ref{tab:decomposition_quality}: (++) corresponds to a human-level decomposition ; (+) corresponds to a satisfying decomposition that a human subject would slightly change ; (-) represents cases where some subquestions are weakly relevant, but will likely not break the reasoning chain ; (- -) corresponds to a false or non-existing decomposition. We choose DeepSeek-R1-Distill-Qwen-32B \cite{deepseek-ai_DeepSeek-R1-Distill-Qwen-32B_2025} for the rest of this study.

\begin{table}[htb!]
    \centering
    \begin{tabular}{lcccc}
        \toprule
        \textbf{Quality} &
        \textbf{++} &
        \textbf{+} &
        \textbf{-} &
        \textbf{- -} \\
        \midrule
        7B/14B/32B &
        27/34/69.5 &
        11/11/12.5 &
        27/24.5/11.5 &
        35/30.5/7.5 \\
        \bottomrule
    \end{tabular}
    \caption{Percentage of decomposition quality per model size across four levels: Excellent (++), Good (+), Weak (-), and Poor (- -). Evaluation was performed on a random sample of 200 questions.}     
    \label{tab:decomposition_quality}
\end{table}

For answer generation, we use LLaMA-2-7B \cite{touvron2023llama} and we use a generation pipeline similar as in \cite{he2024g}. We conduct evaluations on CWQ \cite{talmor2018web} and WebQSP \cite{yih2016value}, two multi-hop QA benchmarks derived from Freebase \cite{bollacker2008freebase}. We follow \cite{luo2023reasoning} for preprocessed datasets and splits. CWQ includes 34,689 questions, while WebQSP includes 4,737 questions. We measure retrieval quality with the following metrics: percentage of connected graphs, average graph density, strong matching, and exact matching. In this study, strong matching measures how often a retrieved subgraph contains a highly relevant node (at least 0.95 similarity with the answer); exact matching measures how often one of the retrieved nodes exactly matches the answer. We measure QA accuracy through Hit@1 metric.\\

Our experiments address two central questions: \textbf{(RQ1)} How does $\alpha$ affects subgraph structure and content? \textbf{(RQ2)} How does this trade-off affects reasoning?

\subsection{The Structure-Content Trade-off (RQ1)}

\begin{figure}[h!]
\centering
\includegraphics[scale=0.45]{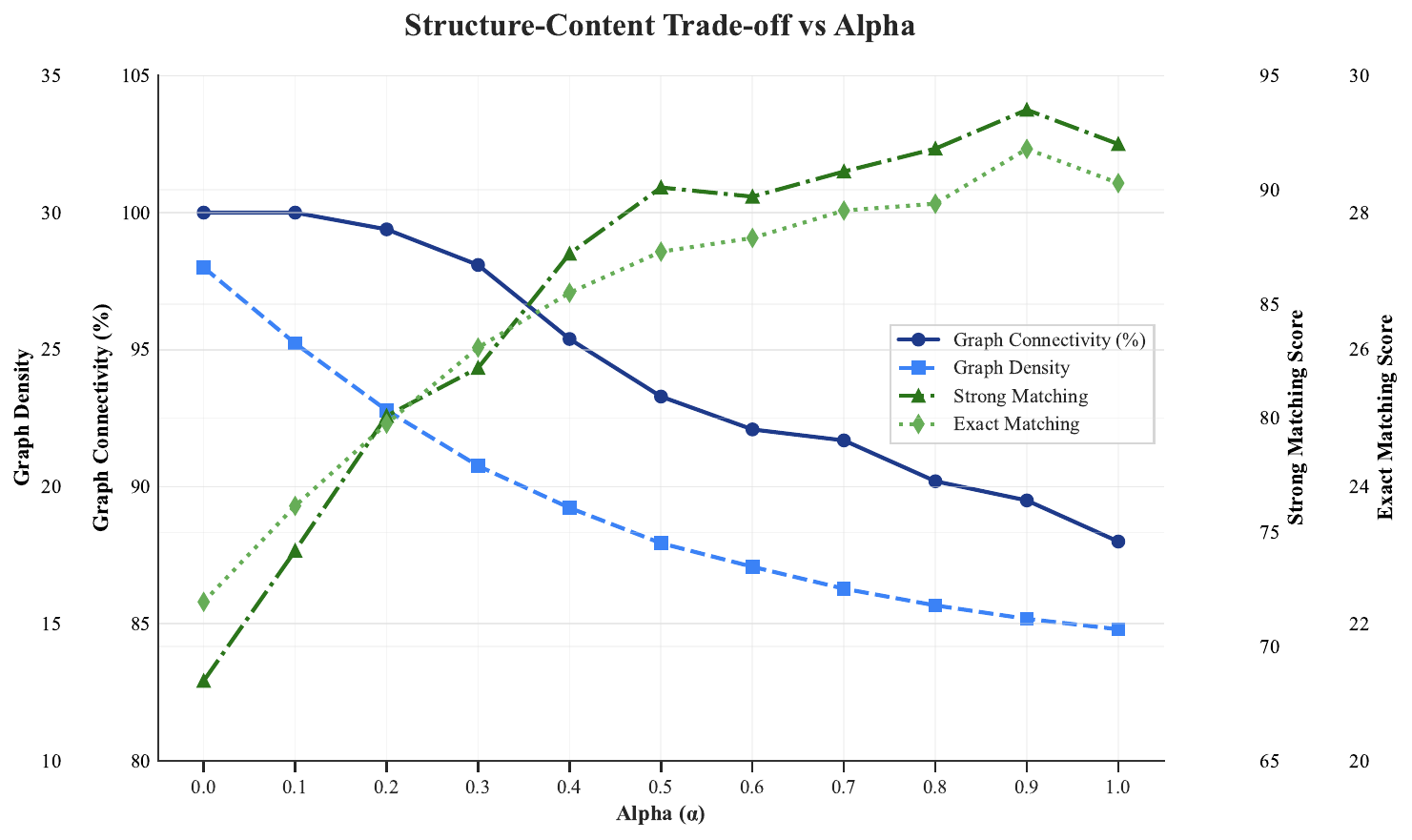}
\caption{Effect of $\alpha$ on subgraph structure (blue) and content (green). Lower $\alpha$ (focusing on initial question) increases connectivity and density, but lowers content relevance; higher $\alpha$ (focusing on subquestions) yields opposite observations.}
\label{fig:graph_metrics}
\end{figure}

\begin{figure}[h!]
\centering
\includegraphics[scale=0.8]{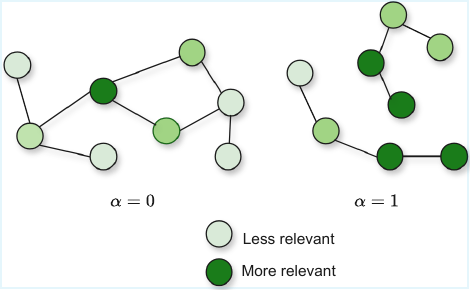}
\caption{Example of retrieved subgraphs for different setups: $\alpha = 0$ (focusing on initial question), and $\alpha = 1$ (focusing on subquestions).}
\label{fig:illustration}
\end{figure}

As shown in Figure~\ref{fig:graph_metrics}, lower $\alpha$ values increase graph connectivity and density but reduce content matching; higher $\alpha$ values produce the opposite trend. Empirically, there seems to be a trade-off between content and structure of the retrieved subgraph. Intuitively, focusing on the initial question forces the retrieval process to focus on a specific region of the knowledge graph to ensure subgraph connectivity; although the resulting subgraph structure is more predictable, the relevance of the retrieved nodes and edges might be limited. On the other side, if we choose to perform multiple retrieval steps using the subquestions, we  maximize relevance of nodes and edges, at the cost of obtaining multiple disjoint subgraphs (see Figure~\ref{fig:illustration}). This leads to the key question: how does this intrinsic trade-off impact LLM answer generation?

\subsection{Impact on Reasoning Performance (RQ2)}

\begin{figure}[h!]
\centering
\includegraphics[scale=0.35]{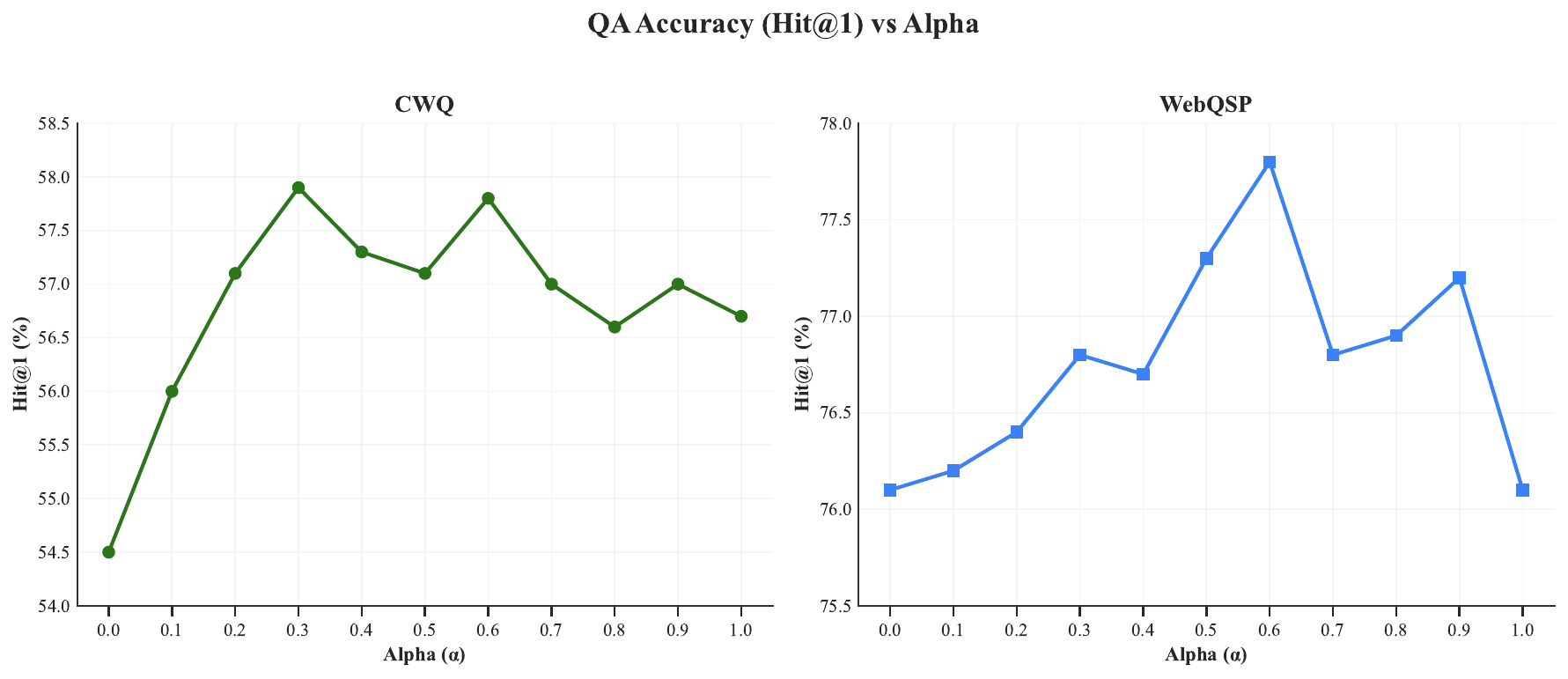}
\caption{QA accuracy peaks at intermediate $\alpha$, showing the benefit of balancing structure and content. Note that $\alpha = 0$ corresponds to the setup in G-Retriever.}
\label{fig:accuracy}
\end{figure}

Figure~\ref{fig:accuracy} shows that reasoning accuracy peaks at intermediate $\alpha$ values (0.3 to 0.7), confirming that none of the extremes yields optimal results. When providing a graph as a prompt to the LLM, the lack of coherent structure makes the content harder to digest: if we provide disjoint graph components in the prompt, the LLM will be missing the connection between these components. If we instead force connectivity, the content will be less relevant on average: this is mostly because we retrieve reasonably small subgraphs to prevent noisy nodes and edges. Therefore, intermediate settings help to balance both graph content and structure, and leads to improved downstream accuracy. This is further supported by analyzing the influence of graph connectivity at $\alpha = 1$: connected subgraphs achieve 55\% Hit@1 versus 48\% for disconnected ones, confirming connectivity's independent contribution to reasoning accuracy.

\section{Conclusion and Future Work} 

We introduce a hybrid retrieval framework, revealing that question decomposition induces a fundamental structure-content trade-off in knowledge graph retrieval. Subquestion-focused retrieval ($\alpha = 1$) yields precise but disconnected subgraphs, while question-focused retrieval ($\alpha = 0$) ensures connectivity at the cost of relevance. QA performance peaks for $\alpha \in [0.3, 0.7]$, where 85-90\% of graphs remain connected while retaining high content precision. These findings show that decomposition improves precision by fragmenting the graph, and that downstream accuracy is maximized when content and structure are jointly balanced, highlighting the need for retrieval systems that adaptively control both aspects. A limitation of our study is the reliance on LLM-generated subquestions, whose variable quality may influence the observed trade-off, as well as the use of fixed parameters ($k=5$, global $\alpha$) that restrict adaptivity. Future work should analyze robustness to propagated decomposition noise and explore learning both the decomposition strategy and $\alpha$ dynamically based on question complexity.\\

\textit{This work has been partially funded by the European Union's Horizon RIA
research and innovation program under grant agreement No. 101189679 (ASTIR). 
This work also benefited from the FactoryIA supercomputer, financially supported by the Ile-de-France Regional Council.}

\begin{footnotesize}
\bibliographystyle{unsrt}
\bibliography{references}
\end{footnotesize}
\end{document}